\UseRawInputEncoding
\documentclass[runningheads]{llncs}
\usepackage{amsmath}
\usepackage[T1]{fontenc}
\usepackage{etoolbox}

\AtBeginEnvironment{thebibliography}{\footnotesize}

\usepackage{graphicx}
\usepackage{booktabs}
\usepackage{tabularx}
\usepackage{multirow} 
\usepackage[accsupp]{axessibility}
\usepackage{color}
\usepackage{orcidlink}
\begin{document}
\title{LSReGen: Large-Scale Regional Generator via Backward Guidance Framework} 
\titlerunning{LSReGen}
%
%
\author{Bowen Zhang\inst{1} \and Cheng Yang\inst{2} \and Xuanhui Liu\inst{2}}
\authorrunning{ B. Zhang et al.}
%
\institute{Zhejiang University  \and Hangzhou University }
\maketitle              

\begin{figure*}[t]
    \centering
    \includegraphics[width=0.9\textwidth]{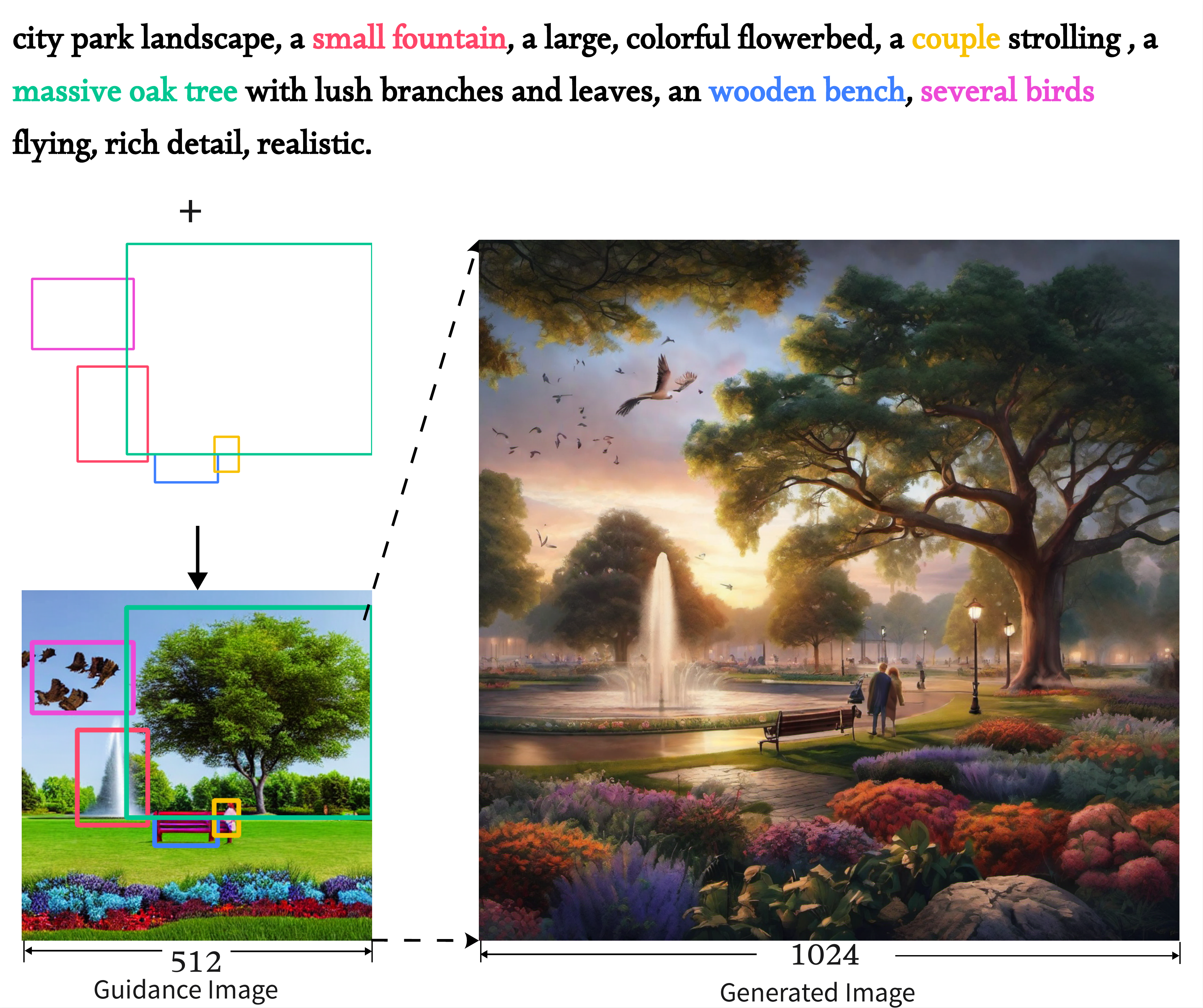}
    \vspace{-.2cm}
\caption{
 {\footnotesize Our approach, LSReGen, takes bounding boxes as input and utilizes the low-parameter pre-trained layout-to-image model GLIGEN as a preprocessor to extract low-frequency information from images. The information serves as layout features to guide the sampling process, resulting in the generation of larger-scale images with richer elements and higher quality.}}
\vspace{-.2cm}
\label{fig:head}
\end{figure*}

\begin{abstract}
In recent years, advancements in AIGC (Artificial Intelligence Generated Content) technology have significantly enhanced the capabilities of large text-to-image models. Despite these improvements, controllable image generation remains a challenge. Current methods, such as training, forward guidance, and backward guidance, have notable limitations. The first two approaches either demand substantial computational resources or produce subpar results. The third approach depends on phenomena specific to certain model architectures, complicating its application to large-scale image generation.
To address these issues, we propose a novel controllable generation framework that offers a generalized interpretation of backward guidance without relying on specific assumptions. Leveraging this framework, we introduce LSReGen, a large-scale layout-to-image method designed to generate high-quality, layout-compliant images. Experimental results show that LSReGen outperforms existing methods in the large-scale layout-to-image task, underscoring the effectiveness of our proposed framework. Our code and models will be open-sourced.
  \keywords{Large-Scale \and Controllable Generation \and Image Editing}
\end{abstract}
\section{Introduction}
\label{sec:intro}
In recent years, AIGC (Artificial Intelligence Generated Content) technology has undergone rapid and significant advancements, achieving a series of remarkable breakthroughs. The advancements in this field have been particularly noteworthy in the field of computer vision, where novel text-to-image methods, such as DALL-E3\cite{betker2023improving}, Imagen\cite{saharia2022photorealistic} and SDXL\cite{podell2023sdxl} provide more sophisticated and innovative means for generating high-quality and large-scale images, even rivaling those of professionals in related domains.

While text-to-image models exhibit powerful generative capabilities, their abilities for high-level control, such as generation of specific concepts, object replacement and layout of a composition, are often limited due to the inadequacy of text in expressing certain complex concepts and constraints imposed by the text encoder's capacity. Rather than focusing on improving the model's textual understanding, incorporating new modalities as a supplementary inputs is a more promising choice. Existing controllable generation methods\cite{zhang2023adding,balaji2022ediffi,brooks2023instructpix2pix} largely concentrate on how to make the model comprehend the newly added modalities.

\begin{figure*}[t]
    \centering
    \includegraphics[width=0.9\textwidth]{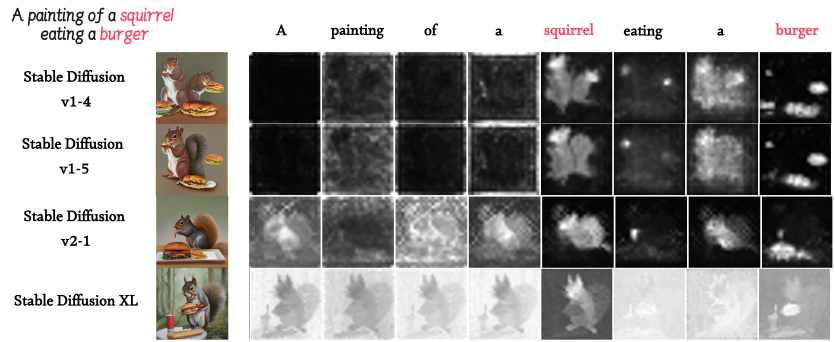}
\caption{The cross-attention maps of different Stable Diffusion versions. In the previous Stable Diffusion models, the attention maps of individual object words only display themselves, while other object words remain hidden in the background. For example, in the attention maps for the word "burger," the squirrel cannot be seen. In contrast, in the SDXL, with a different architecture from previous models, both objects can be clearly seen.}
\vspace{-.2cm}
\label{fig:attention}
\end{figure*}

Prior works on controllable generation can be primarily categorized into three types: model training\cite{yang2023reco,,shen2024boosting,kumari2023multi}, forward guidance\cite{avrahami2023blended,,brack2024sega,hertz2022prompt} and backward guidance\cite{chefer2023attend,mokady2023null,voynov2023sketch,xie2023boxdiff}. Model training involves using datasets that contains additional modalities, fine-tuning\cite{yang2023reco,shen2023triplet,brooks2023instructpix2pix,shen2023advancing} the model globally or locally to imbue it with the ability to understand the semantic content of supplementary modalities, alternatively, training new added modules\cite{li2023gligen,zhang2023adding,ye2023ip} which aim to transform the additional modality into semantic information\cite{shen2023pbsl} comprehensible by the model. The trained model possesses excellent generative control capabilities, but the training incurs significant costs, especially for models with more parameters\cite{podell2023sdxl} and larger-scale datasets. 
Forward guidance encompasses manually modifying the intermediate variables, the noisy images\cite{avrahami2023blended,lugmayr2022repaint} or the attention maps\cite{brack2024sega,tumanyan2023plug,qiao2022novel,li2022enhancing,weng2023cross} during individual sampling steps. This method requires almost no additional overhead, but the generated image results are not satisfactory, such as exhibiting a patchy appearance. Backward guidance method\cite{chefer2023attend,mokady2023null,voynov2023sketch,shen2021exploring,xie2023boxdiff} calculates the distance between the current time step's cross-attention map and the ideal cross-attention map. 
By computing gradients, it updates the noise image for this time step, gradually fitting the intermediate variables of the denoising process to the ideal distribution. This approach strikes a good balance between model training and forward guidance, achieving favorable results with minimal overhead at the inference stage only. 
However, most backward guidance methods rely on the observation\cite{hertz2022prompt} that the values of the cross-attention maps indicate the strength of the influence of individual words on image pixels, thereby reflecting the position of the object in the image. We observed that this phenomenon does not occur for models with more parameters and specific architectures\cite{podell2023sdxl}.
Fig. \ref{fig:attention} compares the cross-attention maps of different version of Stable Diffusion.

\begin{figure*}[t]
    \centering
    \includegraphics[width=0.9\textwidth]{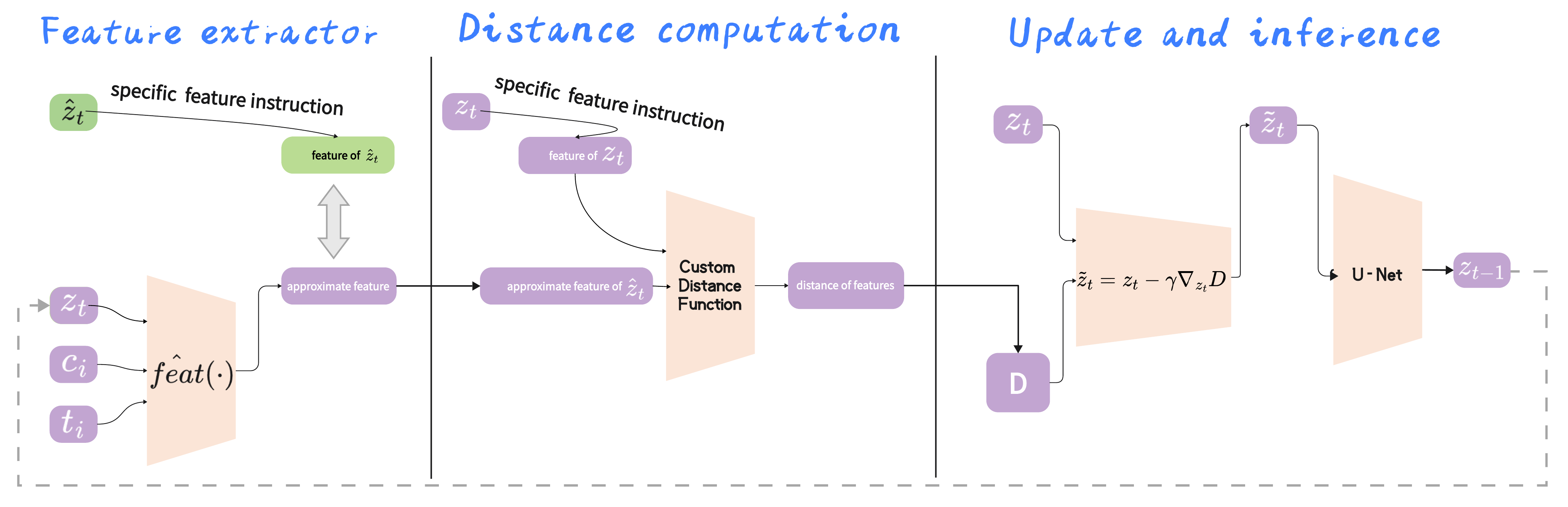}
\caption{Backward guidance framework. Implementing backward guidance during the image sampling process enables controllable image generation. We propose a general interpretation that encompasses backward guidance. The feature extractor extracts approximate features of the ideal image. During sampling, controlling the target is achieved by minimizing the difference between the features of the intermediate variables and the ideal image.}
\vspace{-.2cm}
\label{fig:framework}
\end{figure*}

We provide a universal backward guidance framework that encompasses a general interpretation of backward guidance without relying on cross-attention maps. As shown in Fig. \ref{fig:framework}, the framework takes control information as input, which can be in various forms such as segmentation maps or simple textual instructions. The entire framework requires two custom functions: a feature extraction method and a distance calculation function. The feature extraction method takes in the given control information and, in conjunction with the intermediate variables during generation process, approximates certain features of the ideal intermediate variable. The distance function computes the difference between the "ideal intermediate variable feature" and the "real intermediate variable feature". The framework updates the real intermediate variable by computing the gradient of the difference, guiding it to naturally shift towards the control objective. This process will be repeated throughout the entire or partial sampling steps.

Building upon this framework, we further propose a large-scale layout-to-image method, Fig. \ref{fig:head} illustrates general process of our method. It utilize a pre-trained low-parameter layout-to-image model\cite{li2023gligen} as a feature extractor. By upsampling and adding noise to the small-scale image generated by this model, it captures the layout features of the image. Furthermore, in the generation process of large-scale image, we use the square of the L2 norm to calculate distance between this set of features and the layout features of the intermediate variables. 

Our experiments demonstrate that LSR can generate large-scale images that conform layout information, and the quality of generated images surpasses the existing layout-to-image methods. The success of this method further validates the effectiveness of the framework we proposed. Our contributions are as follows:
\begin{itemize}
    \item We provide a training-free universal backward guidance framework that offers a general interpretation of backward guidance without relying on cross-attention maps.
    \item Based on this framework, we introduce LSReGen, a large-scale layout-to-image method capable of generating high-quality, layout-compliant large-scale images.
    \item Our experimental results demonstrate that LSReGen outperforms existing methods in the large-scale layout-to-image task, underscoring the effectiveness of our proposed framework.
\end{itemize}

\section{Related Work}

\subsection{Text-to-Image Generation}  
With the increase in the scale of data and the emergence of transformers\cite{vaswani2017attention}, approaches based on auto-regressive\cite{ding2021cogview,gafni2022make,yu2022scaling,ramesh2021zero} and diffusion models\cite{nichol2021glide,saharia2022photorealistic,gu2022vector} have gradually become the mainstream focus of research in recent years, which jointly input text encoding and image encoding into a transformer block. They circumvented the shortcomings of GANs, such as model non-convergence and mode collapse. The emerging large text-to-image diffusion models\cite{rombach2022high} encode text using the CLIP text encoder and perform weighted summation on the text encoding for each patch of image through dot product with image encoding, enabling fine-grained text control over the generated images.

To meet the demands of generating large-scale images, \cite{betker2023improving,podell2023sdxl} opted to train models on datasets containing large-scale text-image pairs. Our method bases on SDXL\cite{podell2023sdxl}, which is trained on $1024 \times 1024$ images and modifies the conventional U-Net\cite{ronneberger2015u} architecture by reducing the sampling layers and incorporating additional attention blocks.

\subsection{Controllable Generation in Diffusion Models}  
In order to enhance the controllability of the image generation process, recent approaches\cite{kim2023dense,zhang2023adding,ruiz2023dreambooth,balaji2022ediffi,brooks2023instructpix2pix} tend to incorporate information of additional modalities as inputs to the model. The key challenges lies in facilitating the models understanding of the semantic content introduced by the added modalities.

\subsubsection{Training.}
An intuitive approach to enable the model to understand the semantic content of the additional modalities is to train the model on a dataset that includes the supplementary modality. \cite{yang2023reco,kim2023dense,shen2023git,brooks2023instructpix2pix,kumari2023multi} involve fine-tuning or retraining the model, either globally or locally, to enable the model to comprehend the semantic information from the additional modality without altering the model architecture. \cite{li2023gligen,zhang2023adding,ye2023ip} incorporate additional modules, training on the added modules to transform the supplementary modality into semantic information understandable by the models.
 

\subsubsection{Backward guidance.}
Backward guidance method~\cite{chefer2023attend,mokady2023null,voynov2023sketch,parmar2023zero,xie2023boxdiff,chen2024training,couairon2023zero} designs objective functions targeting the specific requirements, such as layout-to-image\cite{xie2023boxdiff,chen2024training,couairon2023zero} and image-to-image translation\cite{voynov2023sketch,parmar2023zero}, updating the intermediate variables of the denoising process through backpropagation. This approach, compared to forward guidance, enables the generation of more natural images.

\section{Method}
\subsection{Preliminaries: Stable Diffusion}
Stable Diffusion\cite{rombach2022high} is an open-source, state-of-the-art large-scale text-to-image model based on diffusion method\cite{ho2020denoising}. Unlike typical diffusion methods\cite{ho2020denoising,song2020denoising,song2020score}, it operates on the latent space of images.Specifically, Stable Diffusion\cite{rombach2022high} comprises an encoder $E$ and a decoder $D$. For a given image $x$, the encoder compresses the image into a latent variable $z$, and the decoder reconstructs the image from the latent, i.e., $\tilde{x}=D(z)=D(E(x))$. Stable Diffusion introduces noise to the latent using the following formula to obtain the input for the image side during training:$$z_t = \alpha_tz + \sqrt{1 - \alpha_t}\epsilon_t,$$ where $\epsilon_t$ is normally distributed noise and $\alpha_t$ is a decreasing sequence, from $\alpha_0 \approx 1$ to $\alpha_T \approx 0$, representing the noise schedule. As a crucial component, the denoiser $\epsilon_\theta$ of Stable Diffusion employs an modified U-Net architecture, incorporating self-attention layers and cross-attention layers at each sampling block. The pre-trained CLIP's\cite{radford2021learning} text encoder $\tau_\varphi$ receives textual content $y$ to generate preprocessed text encodings $\tau_\varphi(y)$, which are then used as the key and value inputs for the cross-attention layers in the denoiser. The denoiser is responsible for incorporating the text input to predict the noise added during the latent variable's noising process, subsequently calculating the loss via:$$L_{SD}:= \mathbb{E}_{E(x),y,\epsilon\sim{N(0,1)},t}[\Vert\epsilon-\epsilon_\theta(z_t, t, \tau_\varphi(y))\Vert_2^2],$$ where both $\tau_\theta$ and $\epsilon_\theta$ are jointly optimized. During inference, Stable Diffusion samples a random variable $z_T$ from a standard Gaussian distribution $N(0,1)$. It then progressively removes noise to obtain denoised latent $z_0$ using DDPM\cite{ho2020denoising} or other sampling strategies, and finally decodes it using the decoder $D$ to generate the image $\tilde{x}$.



\subsection{Backward Guidance Framework}  
For the image $\hat{x}$ with a special structure, the intermediate variables $\hat{z}_{1...T-1}$ in its generation process also follow a particular distribution, as demonstrated by SDEdit\cite{meng2021sdedit}. In other words, by identifying the distribution characteristics of the ideal intermediate variables, the model can be guided to generate the desired results. Furthermore, for specific control modes, achieving the desired effect can be as simple as extracting the required feature from intermediate variables. For example, in a layout-to-image task, one only needs to focus on the pixel values within the given bounding boxes in $\hat{z}_{1...T-1}$. We sample a random variable $z$ from a Gaussian distribution using it as the image input for the framework. The control information and text information are defined as $ci$ and $ti$, serving as the control input and text input for the framework, respectively. Then, we combine the control information with the random variable through a custom function $\overset{\backsim}{feat(\cdot)}$ to approximate the corresponding features of the ideal variable $\hat{z}_{1...T-1}$, i.e.: $$feat(\hat{z_t})\approx \overset{\backsim}{feat}(z_t, ci, ti).$$ A function $D=distance(feat(z_t), \overset{\backsim}{feat}(z_t, ci, ti))$ that calculates the difference between features needs to be specified, where a smaller function value indicates that the corresponding features of the intermediate variable are closer to those of the ideal variable. We compute the gradient $\nabla_{z_t}D$ of the distance with respect to $z_t$ and update $z_t$ through: $$\tilde{z}_t=z_t - \gamma\nabla_{z_t}D,$$ where $\gamma$ represents the step size used for the update. We use $\tilde{z}$ and $ti$ as inputs to the generator, take the generator's output as the image input for the framework at the next timestep, and repeat this process until the entire sampling procedure of the generator is completed.

\begin{figure*}[tp]
    \centering
    \includegraphics[width=\textwidth]{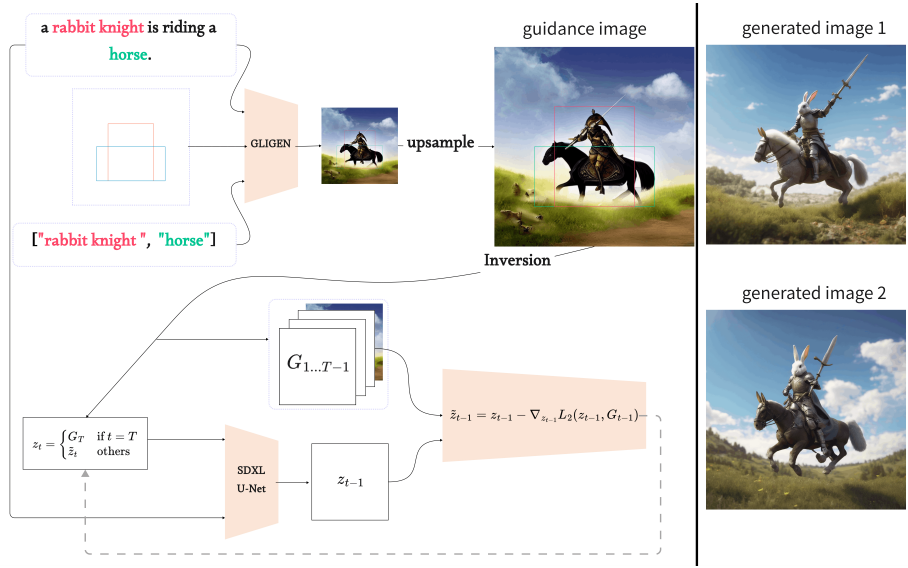}
\caption{Overview of LSReGen. On the left, the general flow of our approach is illustrated. On the right are two different images generated using our approach, but their layouts remain consistent with the provided positional information.}
\vspace{-.2cm}
\label{fig:method}
\end{figure*}

\subsection{Large-Scale Regional Generator}  
With the proposed backward guidance framework, research on controllable generation can be focus on defining feature extraction methods and difference calculation functions. Layout-to-image task takes layout information as input, such as segmentation maps or bounding box coordinates. The goal of the task is to generate an image where the positions of objects are consistent with the given layout.

As confirmed by previous work\cite{kim2023dense,xie2023boxdiff,chen2024training,couairon2023zero}, the low-frequency information $lfi$ in an image or latent reflects the layout of the image. Therefore, we consider it as the "ideal intermediate variable feature" in layout-to-image task. As shown in figure \ref{fig:method}, training-based method, such as GLIGEN\cite{li2023gligen}, show satisfactory performance in layout-to-image task at resolutions of $512 \times 512$. We use these methods to generate small-scale images $x_{small}$ consistent with the given layout and upsample the generated images to target resolution. The encoder $E$ compresses the upscaled image to obtain the last latent variable $\bar{z}_0$. According to the training characteristics of the diffusion method\cite{ho2020denoising} and the perspective proposed in \cite{si2023freeu}, the high-frequency information in the noisy images or latents at large timesteps(close to sampling steps $T$) is essentially eliminated, while low-frequency information is preserved, i.e., $$lfi(z_t) = z_t,  \quad \text{when $t>=tth$},$$ where $tth$ represents a threshold of timesteps. We use the inversion method to obtain $z_{tth...T}$, thus obtaining high-quality low-frequency information, completing the feature extraction.

We use the square of L2 norm, as shown below, as the standard measure of distance: $$distance(lfi(\bar{z}_{t}), lfi(z_t)) = \vert \bar{z}_t - z_t\vert_2^2 \quad t>=tth. $$The guidance framework is used in the early stages of sampling, and afterward, the model executes standard generation steps to ensure the diversity and authenticity of the generated images.   

\section{Experiment and Analysis}
\label{sec:exp}

\begin{figure*}[t]
    \centering
    \includegraphics[width=\textwidth]{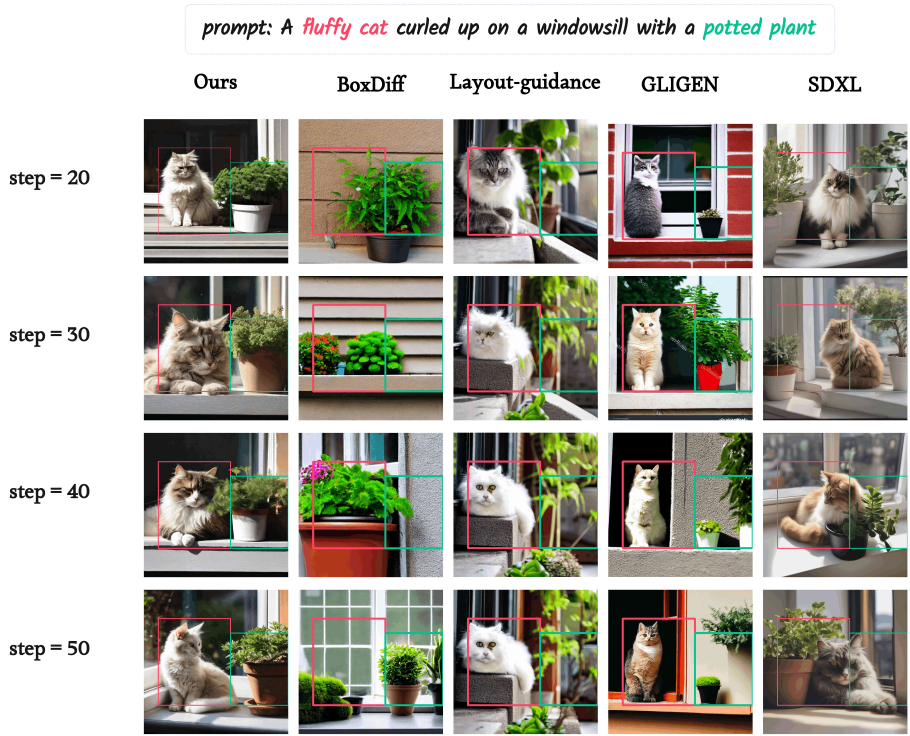}
\caption{Comparison with other methods in different sampling steps. Our method, GLIGEN\cite{li2023gligen} and layout-guidance\cite{chen2024training} are hardly influenced by the number of sampling steps, while BoxDiff\cite{xie2023boxdiff} needs more sampling steps to keep the accuracy of the objects' position.}
\vspace{-.2cm}
\label{fig:comparison}
\end{figure*}

\subsection{Experimental setup}
\textbf{\emph{Datasets.}}
We choose to compare against other works on the COCO validation 2017 dataset, which comprises 5,000 images and bounding boxes for 80 categories. 
Regarding the images in the dataset, we use BLIP2\cite{li2023blip} for text annotation to provide prompts used to generate the visual data. However, providing a reasonable description covering every object in the image using natural language is challenging when there are multiple objects present. Given this scenario, we extracted the two objects with the largest pixel areas in images containing multiple objects and provided the prompt "\textit{what is the relationship between {\emph{object1}} and {\emph{object2}}}" during BLIP2\cite{li2023blip} inference, which object1 and object2 are the two objects we extracted.
 
\textbf{\emph{Evaluation Metrics.}}
We use two standard metrics to evaluate the quality of image generation. Frechet Inception Distance (FID) is used to assess the quality and diversity of the generated images. We used the code "torch-fidelity" to evaluate the 5k generated images. The mean Average Precision(mAP) metric measures the alignment between the objects in the generated images and the provided spatial information. We use the YOLOv5s model to perform object detection on the generated images to calculate mAP. Additionally, we calculate the CLIP score\cite{hessel2021clipscore}, which measures the consistency between the text prompts and the generated images.

\textbf{\emph{Implementation Details.}}
Our method adopts the model architecture of Stable Diffusion(SD) XL\cite{podell2023sdxl}, with the stable-diffusion-xl-base-1.0 model parameters serving as the default pre-trained image generator. The scale of the generated images is three times that of the corresponding image scale in COCO dataset. 
For the two methods used for comparison, BoxDiff\cite{xie2023boxdiff} and layout-guidance\cite{chen2024training}, we adhere to their respective settings. They cannot be applied to SDXL\cite{podell2023sdxl} as both of them leverage the characteristics of cross-attention maps. Additionally, due to the complex gradient computations required by these two methods, increasing the image scale can lead to insufficient memory. We specify these two methods to generate $512 \times 512$ images. In theory, this setting should improve the quality of the generated images, thereby not compromising the persuasiveness of the experiments. All images used in comparison and exhibition are generated on a single NVIDIA A800 GPU.

\begin{figure*}[t]
    \centering
    \includegraphics[width=\textwidth]{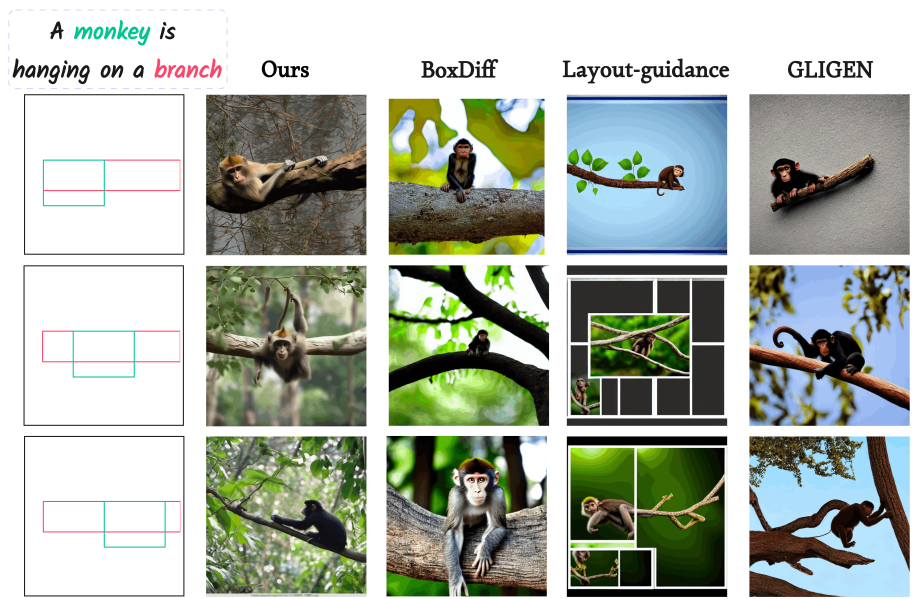}
\caption{Synthesized samples obtained with various location.}
\vspace{-.2cm}
\label{fig:location}
\end{figure*}

\subsection{Comparison with State-of-the-art Methods} 
Figure \ref{fig:comparison} illustrates the quality of image generation and the accuracy of object position for different methods at various sampling steps. It can be observed that both BoxDiff\cite{xie2023boxdiff} and layout-guidance\cite{chen2024training} suffer from the issue of missing objects in the generated images, with BoxDiff\cite{xie2023boxdiff} requiring a larger number of sampling steps to ensure relatively accurate object positions. GLIGEN\cite{li2023gligen} is capable of generating all objects while ensuring the accuracy of their positions, but due to limitations in the model architecture and training data, the quality of the generated images is low. Our method demonstrates the same ability as SDXL to generate high-quality images while ensuring the accuracy of object positions.

\begin{table*}[ht]
\centering 

\label{tab:quantitive} 

\caption{Comparison with other layout-to-image methods. Our method improves spatial fidelity. In mAP, compared to the method of generating $512 \times 512$ images, our approach ranks second only to GLIGEN, but far outperforms it in generating large-scale images. mAP is calculated with an IoU threshold of 0.5. The text-to-image similarity is measured using CLIP score.} 
\vspace{-.2cm}
\begin{tabularx}{0.88\textwidth}{lccccc@{}}
\toprule
\multirow{2}{*}{Method} & \multicolumn{1}{c}{\multirow{2}{*}{FID($\downarrow$)}} & \multicolumn{2}{c}{Single Object} & \multicolumn{2}{c}{Multi Objects} \\ \cmidrule(l){3-6} 
                        & \multicolumn{1}{r}{}                        & mAP($\uparrow$)          & T2I-Sim($\uparrow$)   & mAP($\uparrow$)          & T2I-Sim($\uparrow$)   \\ \midrule
GLIGEN\cite{li2023gligen}                  & 43.0                                        & 89.2            & 30.5            & 65.7            & 29.4            \\ \midrule
GLIGEN(large-scale)\cite{li2023gligen}     & 125.7                                       & 10.6            & 27.4            & 4.1             & 26.0            \\
BoDiff\cite{xie2023boxdiff}                 & 49.5                                        & 17.7            & 29.8            & 6.4             & 28.6            \\
Layout-guidance\cite{chen2024training}         & 47.6                                        & 50.0              & 31.1            & 35.6            & 30.2            \\
SDXL\cite{podell2023sdxl}                    & 46.1                                        & 18.0              & \textbf{30.8}   & 4.5             & 29.6            \\
\textbf{Ours}                    & \textbf{36.6}                               & \textbf{56.2}   & 30.6            & \textbf{36.7}   & \textbf{30.2}   \\ \bottomrule
\end{tabularx}
\smallskip

\end{table*}
\smallskip

In the table \ref{tab:quantitive}, we present the quantitative evaluation results with the state-of-the-art methods. Due to the memory consumption exceeding 80GB for generating large-scale images with BoxDiff\cite{xie2023boxdiff} and Layout-guidance\cite{chen2024training}, we analyze images generated at $512 \times 512$ resolution for these two methods. Setting the resolution to $512 \times 512$ allows them to perform better compared to generating large-scale images. Therefore, our experimental results remain valid. 
It can be observed that due to the use of a more powerful pre-trained generative model, our approach can generate images with higher fidelity and diversity. Compared to the SDXL model, our approach also achieves a leading position. In terms of layout consistency, our method achieves the highest level on large-scale images. While GLIGEN performs well on 512x512 images, it lacks the ability to ensure layout consistency on large-scale images. In terms of text-to-image similarity, our method maintains the performance of the underlying model without degradation. Compared to all existing layout-to-image methods, it demonstrates superior capabilities.

Fig. \ref{fig:location} presents synthesized samples using different methods with varying bounding boxes. The ability of the image structure to vary with the same prompt and a varying bounding box is not sufficiently strong in BoxDiff\cite{xie2023boxdiff} and layout-guidance\cite{chen2024training}. While GLIGEN\cite{li2023gligen} is able to capture changes in position, the fidelity of the image is somewhat reduced. Our method ensures that the object follow the position of the bounding box without compromising fidelity, demonstrating strong robustness.
 
\begin{figure}[ht]
    \begin{minipage}[t]{0.45\textwidth}
    \centering
    \includegraphics[width=\linewidth]{tau_comparison.pdf}
    \caption{The intermediate variables in the sampling process with different numbers of guiding steps. The layout are already evident in the early stages of the sampling process, and excessive guiding steps may instead lead to a decrease in the quality of the generated images.}
    \vspace{-.2cm}
    \label{fig:tau_comparison}
    \end{minipage}
    \hfill
    \begin{minipage}[t]{0.45\textwidth}
    \centering
    \includegraphics[width=\linewidth]{scale_comparison.pdf}
    \caption{Qualitative comparison of different loss scales in the backward guidance. The scale are increased from left to right keeping the same prompt. The higher scale, the more tightly objects are constrained inside the bounding boxes. However, for very high scales, fidelity decreases significantly.}
    \vspace{-.2cm}
    \label{fig:scale_comparison}
    \end{minipage}

\end{figure}

\subsection{Ablation Studies and Analysis}

\textbf{\emph{Percentage of Guidance Steps.}}
We investigated the appropriate steps that balance image quality and layout control. The intermediate variables generated during the image generation process using different percentages of guidance steps are illustrated in Fig. \ref{fig:tau_comparison}. We found that the layout information of the image emerged as early as the second step in the sampling process, which corresponds to the first 10\% of the steps. If more guidance steps are taken, it actually leads to a decrease in the quality of the generated images. Based on our experiments, we adopt 0.1 as the default percentage of the guidance steps.

\textbf{\emph{Scale Factor of Guidance.}}
In Fig. \ref{fig:scale_comparison}, we qualitatively analyze the impact of guidance scale factor. We found that as the loss weight increases, the model's control over the image also strengthens, but at the cost of reduced fidelity, particularly with higher scales. In practical applications, the guidance strength value should be adjusted based on the size of the input bounding box to achieve the optimal results. For larger bounding boxes, only a small weight is needed to achieve good control effects, while for relatively smaller bounding boxes, the guidance strength needs to be increased.
\label{tab:preference}
\begin{table}[]
\centering
\caption{\textbf{User study.} This study compares our method with two different baselines, BoxDiff and layout-guidance, focusing on image reality, layout alignment, and harmonization. Participants were asked to rank each method based on their preferences.}
\vspace{-.2cm}
\begin{tabular}{l|ccc}
\toprule
Method     &Visual Fidelity {$\uparrow$}   &Consistency {$\uparrow$}  & User Preference {$\uparrow$} \\ \hline
Ours over BoxDiff         & \textbf{86.7\%}  &\textbf{95.0\%} & \textbf{85.0\%}                   \\
Ours over layout-guidance & \textbf{66.7\%}   & \textbf{87.0\%} & \textbf{75.0\%}                    \\ \bottomrule
\end{tabular}
\end{table}
\vspace{-0.3cm}
\subsection{User Study}
In order to align with human preferences, we conducted a comprehensive user study to compare our method LSReGen against the top-performing training-free baselines: BoxDiff\cite{xie2023boxdiff} and layout-guidance\cite{chen2024training}. We recruited 40 participants, divided into 4 groups of 10 users each, to complete 20 evaluation tasks, divided into 2 sets of 10 tasks for each baseline comparison. This resulted in a total of 400 user evaluations. Participants were asked to rank their preferred methods based on three criteria: the realism of the images, consistency with the given layout, and realistic integration of the subject into the background image. The results, presented in Table \ref{tab:preference}, indicate a strong user preference for our generated outputs compared to the baselines.

\section{Conclusion}\label{sec:con}

In this paper, we proposed a universal framework based on backward guidance and explored the potential of layout control on large-scale images using a large pre-trained text-to-image model without training or fine-tuning. We confirmed that the previously observed characteristics of cross-attention maps are not applicable to specific model architectures. Instead, we adopted a pre-trained low-parameter model as a feature extractor. Our experiments have demonstrated the superior performance of this method in layout-to-image tasks, particularly on large-scale images. Finally, we discussed the limitations of this method and the broader domains where the framework could be applied.

\begingroup
\footnotesize
\bibliographystyle{splncs04}
\bibliography{ref}
\endgroup
\end{document}